\documentclass[a4paper,fleqn]{cas-dc}
\usepackage{float}
\usepackage{algorithm}
\usepackage{algpseudocode}
\usepackage{caption}
\usepackage[authoryear,round]{natbib}

\newcommand{\onpokd}{\textsc{OnPoKD}}

\newcommand{\ours}{\onpokd}
\newcommand{\KL}{\mathrm{KL}}
\newcommand{\softmax}{\operatorname{softmax}}
\newcommand{\argmax}{\operatorname{argmax}}
\newcommand{\indicator}{\mathbb{I}}
\newcommand{\clip}{\textsc{CLIP}}
\definecolor{customred}{RGB}{192,0,0}

\ExplSyntaxOn
\RenewDocumentCommand \printorcid { }
{
  \seq_if_empty:NF \g_stm_orcid_seq
  {
    \group_begin:
      \tex_let:D \thefootnote \relax \footnotetext
      {
        \raggedright
        \textsc{orcid}(s):\c_space_token
        \seq_use:Nn \g_stm_orcid_seq { ;~ }
      }
    \group_end:
  }
}
\ExplSyntaxOff

\begin{document}
\let\WriteBookmarks\relax
\def\floatpagepagefraction{1}
\def\textpagefraction{.001}

\shorttitle{On-Policy Distillation for Vision-Language Model Adaptation}
\shortauthors{H. Zhang et al.}

\title [mode = title]{On-Policy Distillation for Vision-Language Model Adaptation, an Effective Paradigm on Low-Quality Multimodal Data}

\author[1,2]{Hongyuan Zhang}
\fnmark[$\dagger$]
\ead{hyzhang98@gmail.com }

\author[3]{Xianda Guo}
\fnmark[$\dagger$,$\ddagger$]
\ead{xianda_guo@163.com}

\author[2]{Yanlun Peng}
\fnmark[$\dagger$]
\ead{allenpeng0209@gmail.com}

\author[2,4]{Qianlong Yang}

\author[5]{Yubin Guo}

\author[2,3]{Pinhan Fu}

\author[6]{Mulin Chen}

\author[7]{Xiaozhen Qiao}
\cormark[1]
\ead{xiaozhennnqiao@mail.ustc.edu.cn}

\author[1]{Ping Luo}
\cormark[1]

\affiliation[1]{organization={The University of Hong Kong},
            country={Hong Kong SAR, China}}
\affiliation[2]{organization={Great Wall Motor},
            country={China}}
\affiliation[3]{organization={School of Computer Science, Wuhan University},
            country={China}}
\affiliation[4]{organization={School of Science, China University of Petroleum (East China)},
            country={China}}
\affiliation[5]{organization={School of Computer Science and Technology, University of Science and Technology of China},
            country={China}}
\affiliation[6]{organization={School of Artificial Intelligence, Optics and Electronics (iOPEN), Northwestern Polytechnical University},
            country={China}}
\affiliation[7]{organization={School of Information Science and Technology, University of Science and Technology of China},
            country={China}}

\cortext[1]{Corresponding author.}
\nonumnote{$\dagger$~These authors contributed equally to this work.}
\nonumnote{$\ddagger$~Project lead.}

\begin{abstract}
Knowledge distillation offers an efficient route to transfer a task-adapted vision-language teacher to a compact student. The training target in current vision-language distillation methods is typically constructed from the teacher prediction and applied uniformly to all training samples, making it unreliable under class and domain shifts. In this paper, we argue that distillation target construction should be treated as a dynamic training decision rather than a fixed recipe. To this end, we propose \ours{}, an on-policy distillation framework for vision-language model adaptation. To the best of our knowledge, \ours{} is the first framework that applies on-policy distillation to vision-language model adaptation by learning target construction as a policy decision. \ours{} learns a lightweight controller that constructs sample-wise adaptive targets using reliability and disagreement cues from the teacher model, student model, and zero-shot prior. Instead of relying on a fixed teacher prediction, the controller dynamically balances teacher supervision, zero-shot prior guidance, and hard-label anchoring through bounded policy actions, allowing the distillation target to adapt to varying sample reliability and training stages. The policy controller is updated with validation feedback, encouraging target construction to optimize transferability rather than merely fitting the training distribution. Since the controller is only used during training, \ours{} can be seamlessly integrated into existing vision-language distillation pipelines while preserving the original inference architecture and test-time cost. Extensive experiments on Base-to-novel generalization and Cross-dataset transfer benchmarks show that \ours{} consistently improves over strong vision-language distillation baselines.
\end{abstract}


\begin{keywords}
Vision-language models \sep Knowledge distillation \sep On-policy distillation
\end{keywords}

\maketitle

\section{Introduction}

Large-scale vision-language models have become a strong foundation for open-vocabulary visual recognition. By aligning images with natural-language supervision, models such as \clip{} can recognize categories from textual descriptions and transfer to new recognition tasks without training a task-specific classifier~\citep{radford2021learning,jia2021scaling,zhai2022lit}. This property is especially valuable when labeled data are limited or when test classes and domains differ from the original training distribution. In practice, however, downstream applications~\citep{qiao2026ahap,zhu2025viewmask,wang2026scene,xu2026scaseg,zhou2026diffusion} still require adaptation, since target datasets often contain fine-grained categories, domain-specific visual statistics, or label spaces that are not well covered by web-scale pre-training data~\citep{recht2019imagenet,hendrycks2021many}. The central challenge is therefore to improve target-task discrimination while preserving the transferable zero-shot prior that supports novel-class and shifted-domain recognition.

\begin{figure}[pos=h,width=\columnwidth]
    \centering
    \includegraphics[width=\columnwidth]{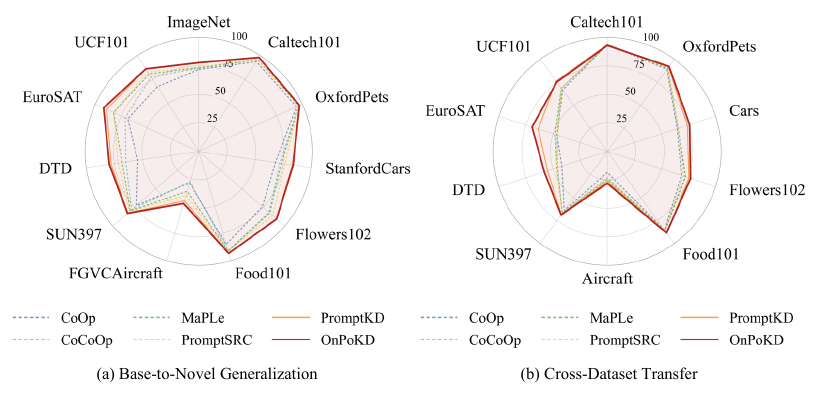}
    \caption{Comparison of \ours{} with existing VLM adaptation methods. (a) Base-to-novel generalization harmonic mean (HM). (b) Cross-dataset transfer accuracy.}
    \label{fig:radar_summary}
\end{figure}

Parameter-efficient adaptation has become a common way to specialize vision-language models while keeping most of the pretrained backbone unchanged. Prompt-based methods replace hand-crafted text templates with learnable contexts~\citep{zhou2022learning,VPN,PiNDA,PiNI,qiao2026semantic}, and later extend them with instance-conditioned or multi-modal prompt learning~\citep{zhou2022conditional,khattak2023maple,khattak2023self,MuNG}. Other methods introduce lightweight adapters or cache-based modules on top of frozen vision-language representations~\citep{yu2023task,lee2023read,qiao2026bidirectional,qiao2025class}. These approaches improve discrimination on adapted base classes, but they also reveal a trade-off in vision-language adaptation. The signals that sharpen source-class decisions do not always preserve the open-vocabulary knowledge needed for novel classes or shifted domains. As adaptation becomes stronger, the model may rely heavily on source-specific visual and semantic patterns, whereas the frozen zero-shot prior can remain a more reliable guide when the test distribution moves away from the adapted training split.

\begin{figure*}[t]
    \centering
    \includegraphics[width=1.00\textwidth]{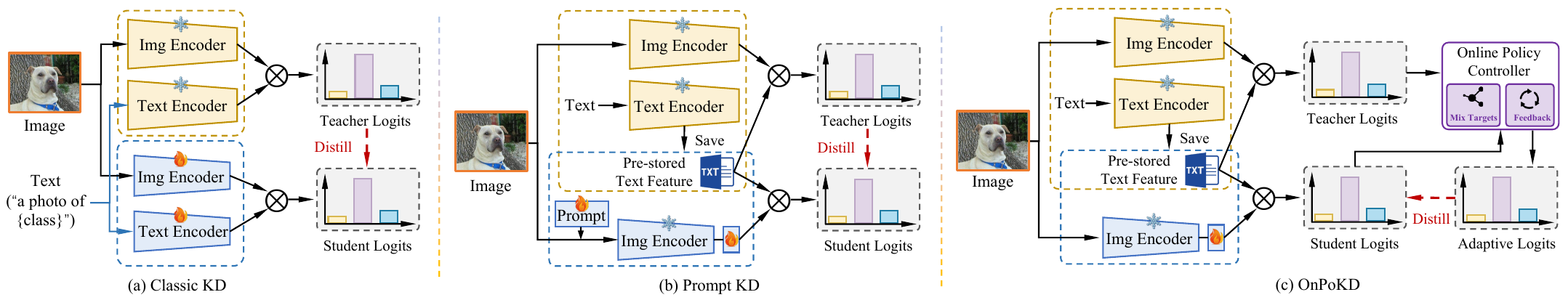}
    \caption{Comparison of different distillation paradigms for vision-language adaptation. (a) Vanilla VLM distillation matches the student to a fixed teacher target. (b) PromptKD distills from a stronger adapted teacher. (c) \ours{} uses an on-policy controller to build a sample-wise adaptive target.}
    \label{fig:overview}
\end{figure*}

Knowledge distillation offers a natural way to transfer adapted recognition ability into a compact student~\citep{hinton2015distilling}. A strong adapted teacher encodes task-specific improvements, and the student inherits them while keeping a simple inference path. Existing vision-language distillation methods build the training target directly from the teacher prediction and optimize the student to match that soft distribution~\citep{wu2023tinyclip,li2024promptkd,yang2024clip}. This design is sound only when the teacher is reliable for every sample, an assumption that rarely holds in vision-language adaptation. The adapted teacher may be accurate on base classes yet unreliable for novel classes, shifted domains, or ambiguous samples, whereas the frozen zero-shot prior, though weaker on the source split, often preserves open-vocabulary information that should not be overwritten. The student itself also changes during optimization, so the target that is best early in training may no longer be best once the student stabilizes. A fixed teacher-centered target cannot express any of these sample-wise and stage-wise differences.

These observations suggest that distillation target construction should be adaptive rather than fixed. The teacher, zero-shot prior, and hard label provide complementary supervision for different training conditions. The teacher transfers task-adapted knowledge, the zero-shot prior preserves the generalization ability of the foundation model, and the hard label anchors learning when soft predictions are uncertain or conflicting. Their relative importance should depend on sample reliability, model disagreement, and training progress. For example, teacher-dominant supervision can be useful at the beginning of optimization, whereas stronger prior or label intervention may become beneficial when the teacher is uncertain, conflicts with the prior, or the student has become more stable. These factors motivate an online mechanism that observes the current training state and constructs a sample-wise distillation target.

We instantiate this mechanism as \ours{}, an on-policy distillation framework for vision-language model adaptation (Figure~\ref{fig:overview}). \textbf{Rather than fixing the mixture between teacher, prior, and label supervision by hand, \ours{} learns a lightweight policy controller during training.} The controller summarizes reliability and disagreement cues from the adapted teacher, the student, and the frozen zero-shot prior, and emits bounded actions for target mixing, sample weighting, and temperature adjustment. Crucially, \textbf{the policy controller is updated from validation feedback}, which steers target construction toward improving transfer behavior instead of merely reducing the current training loss. Because the controller and prior branch are discarded after distillation, the student keeps the same inference architecture and test-time cost as the underlying vision-language adaptation pipeline.
Experiments confirm that adaptive target construction improves vision-language distillation under both class shift and dataset shift. Across eleven Base-to-novel generalization benchmarks, \ours{} raises the average harmonic mean over PromptKD from 83.73 to 84.62, with a larger gain on novel classes than on base classes. In Cross-dataset transfer from ImageNet to ten target datasets, it reaches the best average accuracy of 72.66, improving PromptKD by 1.33\%. Ablation studies further show that a fixed teacher-prior-label mixture is not sufficient, and that reliability cues, validation feedback, and bounded policy actions are all needed for stable target construction. 

We summarize our contributions below.
\begin{itemize}
    \item We are the first to formulate target construction for vision-language distillation as an on-policy decision, enabling sample-wise and stage-wise adaptive supervision.
    \item A lightweight policy controller balances task-adapted teacher, zero-shot prior, and hard-label supervision. This provides more reliable sample-wise targets under class and domain shifts.
    \item We introduce a validation-guided policy update that learns target-construction behavior from reliability feedback, without modifying the teacher, student, or inference architecture.
    \item We demonstrate on Base-to-novel generalization and Cross-dataset transfer benchmarks that \ours{} improves both robustness and transferability.
\end{itemize}

\section{Related Work}

\subsection{Vision-Language Models}

Vision-language models learn transferable visual representations by aligning images with natural-language descriptions~\citep{radford2021learning,jia2021scaling,yao2021filip,yuan2021florence}. They recognize unseen categories through text prompts and provide a strong starting point for downstream recognition with limited labels, but their performance hinges on prompt design and degrades under downstream distribution shift, which makes efficient adaptation important for practical deployment. Prompt learning addresses this issue by replacing hand-crafted prompts with learnable context vectors. CoOp optimizes continuous prompts for downstream classes, while CoCoOp conditions prompts on image features to improve novel-class generalization~\citep{zhou2022learning,zhou2022conditional}. MaPLe extends prompt learning to both the visual and textual branches, and PromptSRC regularizes prompts to preserve the frozen foundation-model representation~\citep{khattak2023maple,khattak2023self}. A complementary line adds lightweight feature adapters or cache mechanisms on top of frozen vision-language representations~\citep{yu2023task,lee2023read}. These studies establish that parameter-efficient adaptation works well for recognition, yet they focus on how to adapt the representation or classifier and leave open how the distillation target should change during student training. \ours{} is orthogonal in this respect. Given an adapted teacher and a student, it improves the training target itself.

\subsection{Knowledge Distillation}

Knowledge distillation transfers information from a stronger teacher to a smaller or more deployable student, usually by matching softened predictive distributions, and has been widely applied to classification, detection, and model compression~\citep{hinton2015distilling,zhao2022decoupled}. It is especially attractive in vision-language adaptation, where a strong adapted teacher may be costly to deploy but a distilled student can retain a compact inference path. PromptKD applies distillation to prompt-based vision-language models and shows that a teacher can improve student prompt adaptation~\citep{li2024promptkd}, while recent \clip{} distillation studies stress the need to preserve cross-modal transfer during compression or adaptation~\citep{wu2023tinyclip,yang2024clip}. Standard distillation, however, assumes the teacher distribution is the right target for every sample. That assumption breaks down when the teacher is unreliable across class splits or domains, and when the frozen zero-shot prior still holds useful information that the adapted teacher has weakened. A single teacher-centered target can neither express these reliability changes nor decide when hard labels should override uncertain soft targets. \ours{} instead treats the target itself as adaptive, keeping teacher-dominant supervision when the teacher is reliable and borrowing the frozen zero-shot prior or hard-label guidance when soft targets are not.

\subsection{Adaptive Policy Learning for Distillation}

Several learning paradigms adjust supervision during training, including sample weighting, curriculum learning, confidence filtering, and teacher-student agreement~\citep{li2023curriculum,yang2021knowledge,zhao2022decoupled}. Related ideas appear in large language models. RLHF updates a language-model policy from human preference feedback~\citep{ouyang2022training}, DPO turns preference optimization into a direct classification-style objective~\citep{rafailov2023direct}, and language-model distillation shows that teacher outputs, rationales, or on-policy trajectories can guide smaller or weaker models~\citep{hsieh2023distilling,zhao2026selfdistilled}. Collectively, these works suggest that supervision can be policy-dependent and that feedback can decide which signals should guide learning. \ours{} adopts this high-level view but targets a different problem. Instead of optimizing a generative language policy, it learns a training-time policy for vision-language distillation target construction that decides which source should dominate, how strongly each sample should be weighted, and how soft the target should be. The controller turns reliability cues from the teacher, student, and prior into bounded actions for target mixing, sample weighting, and temperature adjustment, and is updated by a supervised action-matching step from held-out validation reliability rather than by reinforcement learning or a meta-gradient through the student optimizer. \ours{} is therefore a training-time target-construction mechanism, not a test-time ensemble or an architecture modification.

\begin{figure*}[t]
    \centering
    \includegraphics[width=1.00\textwidth]{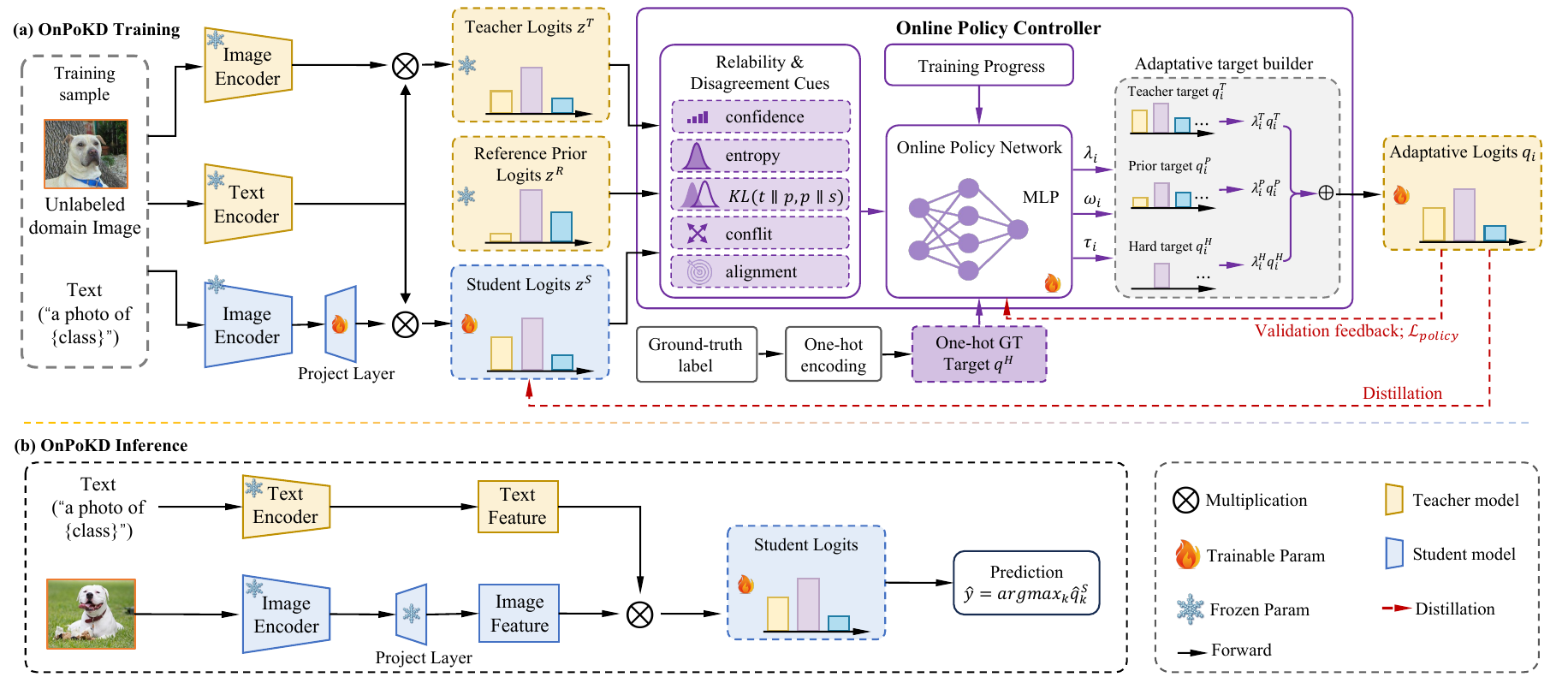}
    \caption{Pipeline of \ours{} for on-policy distillation in vision-language adaptation. During training, a lightweight controller uses reliability cues from the adapted teacher, the student, and the frozen zero-shot prior, together with training progress, to construct adaptive distillation targets. At inference time, the controller and prior branch are removed, leaving the same student architecture and test-time cost.}
    \label{fig:pipeline}
\end{figure*}

\section{Method}

\subsection{Problem Formulation}

Knowledge distillation adapts a compact vision-language model from a stronger teacher, but conventional distillation applies the same teacher-centered target to every training image. This static rule is restrictive in vision-language adaptation, where an adapted teacher may be reliable on base categories yet less stable under class shift, a frozen zero-shot model may preserve useful open-vocabulary priors, and the student may need different supervision as optimization progresses. Together these observations motivate a training mechanism that chooses the supervision source for each sample and training stage.

Let $x_i$ be an input image and $y_i$ be its class label. We consider a teacher-student adaptation setting. It contains an adapted teacher $T$, a student $S$, and a frozen zero-shot vision-language prior $P$. For sample $x_i$, the three models produce logits $z_i^T$, $z_i^S$, and $z_i^P$. Standard logit distillation minimizes
\begin{equation}
    \mathcal{L}_{\mathrm{KD}}(i)
    =
    \tau^2 \KL\left(
    \softmax(z_i^T / \tau)
    \,\middle\|\,
    \softmax(z_i^S / \tau)
    \right),
\end{equation}
where $\tau$ is the distillation temperature and the teacher distribution is the only soft target. This is a forward KL with the teacher as the reference distribution, which is mass-covering, or mean-seeking. It drives the student to place probability mass on every class the teacher deems plausible, including classes where the adapted teacher is wrong under class or domain shift. Because a teacher-centered target has no mechanism to discount these errors, the student imitates the teacher even when the teacher conflicts with the zero-shot prior, and even when stronger label guidance would be more useful. This limitation motivates mixing the teacher prediction with the zero-shot prior and the hard label, two sources that can recover probability mass from teacher-erroneous classes.

\ours{} addresses this limitation by learning sample-wise target construction within an on-policy distillation framework. As shown in Figure~\ref{fig:pipeline}, the on-policy controller summarizes reliability and disagreement cues from the adapted teacher, the student, and the frozen zero-shot prior, and predicts bounded actions for target mixing, sample weighting, and temperature adjustment (Section~3.2). These actions define a policy-guided distillation objective that balances teacher supervision, zero-shot prior guidance, and hard-label anchoring (Section~3.3). The controller is updated with held-out validation feedback to favor transfer-oriented target construction over short-term training-loss reduction (Section~3.4). After training, the controller and prior branch are discarded, leaving the student with unchanged inference architecture and test-time cost (Section~3.5).

\subsection{On-Policy Controller}

The controller captures reliability and conflict among the three supervision sources without adding any extra model evaluations. It first builds a state vector $s_i$ from statistics already available in the forward pass.
\begin{equation}
\begin{aligned}
s_i = [&c_T, m_T, h_T,
c_S, m_S, h_S,\\
&d_{TS}^{\KL}, d_{TS}^{\argmax}, a_{TS},
d_{TP}^{\argmax}, d_{TP}^{\KL}].
\end{aligned}
\end{equation}
Here $c$, $m$, and $h$ denote confidence, top-two margin, and normalized entropy. For a distribution $p_i^M=\softmax(z_i^M)$ from model $M$, we compute
\begin{equation}
\begin{aligned}
c_M &= \max_y p_i^M(y),\\
m_M &= p_i^M(y_1)-p_i^M(y_2),\\
h_M &= -\frac{1}{\log C}\sum_{y=1}^{C}p_i^M(y)\log p_i^M(y),
\end{aligned}
\end{equation}
where $y_1$ and $y_2$ are the top two predicted classes and $C$ is the number of classes. The superscripts $\KL$ and $\argmax$ denote KL divergence and top-label disagreement, respectively, so $d_{TS}^{\KL}$ and $d_{TS}^{\argmax}$ measure teacher-student disagreement while $d_{TP}^{\argmax}$ and $d_{TP}^{\KL}$ measure teacher-prior conflict.
\begin{equation}
\begin{aligned}
d_{AB}^{\KL} &= \KL(p_i^A \| p_i^B),\\
d_{AB}^{\argmax} &= \indicator\!\left[\argmax_y p_i^A(y)\neq \argmax_y p_i^B(y)\right].
\end{aligned}
\end{equation}
The alignment term $a_{TS}$ is the normalized similarity between teacher and student image features. Together these quantities form the controller state, summarizing teacher reliability, student uncertainty, feature transfer quality, and conflict between adapted and zero-shot knowledge.

Training progress acts as an additional policy condition. Rather than concatenating it to the state vector, we use the normalized progress $p\in[0,1]$ to modulate the action logits, which keeps the policy conservative at the start of training and gradually admits stronger prior or hard-label intervention as the student stabilizes.

Given $s_i$ and the normalized training progress $p_i$, the controller first predicts raw action scores
\begin{equation}
    (r_i^\lambda,r_i^w,r_i^\tau)=f_\theta(s_i),
    \label{eq:policy-raw}
\end{equation}
where $r_i^\lambda\in\mathbb{R}^{3}$ controls target mixing. The scalars $r_i^w,r_i^\tau\in\mathbb{R}$ control sample reweighting and temperature. These raw scores are then converted into a policy action
\begin{equation}
    a_i = \pi_\theta(s_i,p_i)
    =
    \left(\lambda_i,w_i,\tau_i\right).
\end{equation}
Here $\lambda_i$ is a three-way mixture over the supervision sources, $w_i$ is a sample weight, and $\tau_i$ is a distillation temperature. We implement $f_\theta$ as a two-layer multilayer perceptron with layer normalization whose final layer is initialized to zero, so the initial behavior follows a conservative teacher-dominant prior action rather than random policy outputs.

\subsection{Policy Guided Distillation Objective}

The controller maps each sample to three-way mixture weights, which we combine with a conservative teacher-dominant base action and with progress-aware adjustment terms. These adjustment terms estimate, from teacher uncertainty, teacher-prior conflict, student uncertainty, and training progress, whether the zero-shot prior or the hard-label target should receive more mass. The resulting mixture action satisfies
\begin{equation}
\begin{aligned}
\eta_i^P &= \phi_P(h_T,d_{TP}^{\argmax},d_{TP}^{\KL}),\\
\eta_i^H &= \phi_H(h_T,d_{TS}^{\argmax},h_S),
\end{aligned}
\label{eq:need-scores}
\end{equation}
where $\eta_i^P$ and $\eta_i^H$ are normalized need scores for the zero-shot prior and hard-label supervision, and $\phi_P,\phi_H$ are monotone summaries of the reliability cues. Prior intervention thus grows when the teacher is uncertain or conflicts with the zero-shot prior, while hard-label intervention grows when the teacher-student decision gap or the student uncertainty is large.

Let $b$ denote a teacher-dominant base mixture and let $\beta_P,\beta_H$ be nonnegative scaling coefficients. The policy mixture is computed as
\begin{equation}
\begin{aligned}
\tilde{\lambda}_i
&=
\softmax\!\left(
\log b
+r_i^\lambda
+[0,\beta_P p_i\eta_i^P,\beta_H p_i\eta_i^H]
\right),\\
\lambda_i
&=\Pi_{\mathcal{C}}(\tilde{\lambda}_i),
\quad
\lambda_i=(\lambda_i^T,\lambda_i^P,\lambda_i^H),
\end{aligned}
\label{eq:policy-mixture}
\end{equation}
where $\lambda_i$ contains the teacher, prior, and hard-label mixture weights. The feasible set
\begin{equation}
\begin{aligned}
\mathcal{C}
=
\{\,\lambda\in\mathbb{R}_{+}^{3}\mid
&\lambda^T+\lambda^P+\lambda^H=1,\\
&\lambda^P\leq \lambda_{\max}^P,\ 
\lambda^H\leq \lambda_{\max}^H,\\
&\lambda^P+\lambda^H\leq \lambda_{\max}^{A}\,\}
\end{aligned}
\label{eq:cap-set}
\end{equation}
keeps the prior and hard-label components as auxiliary supervision sources, and $\Pi_{\mathcal{C}}$ denotes the corresponding cap-and-renormalize projection.

The scalar action components are also bounded.
\begin{equation}
\begin{aligned}
w_i &=
\operatorname{clip}_{[w_{\min},w_{\max}]}
\left(1+\gamma_w(\sigma(r_i^w)-1/2)\right),\\
\tau_i &=
\operatorname{clip}_{[\tau_{\min},\tau_{\max}]}
\left(\tau_0(1+\gamma_\tau(\sigma(r_i^\tau)-1/2))+\delta_\tau(h_T)\right).
\end{aligned}
\label{eq:weight-temp}
\end{equation}
Here $\sigma(\cdot)$ is the sigmoid function, $\tau_0$ is the base distillation temperature, and $\delta_\tau$ is an uncertainty-dependent residual. This parameterization lets the controller adapt sample importance and target softness, while the bounds prevent unstable target construction.

The adaptive target is
\begin{equation}
    q_i =
    \lambda_i^T \softmax(z_i^T / \tau_i)
    +
    \lambda_i^P q_i^P
    +
    \lambda_i^H q_i^H,
    \label{eq:adaptive-target}
\end{equation}
where $q_i^P$ is the prior component and $q_i^H$ is the one-hot label distribution, optionally with label smoothing. When all classes are adapted jointly, $q_i^P=\softmax(z_i^P/\tau_i)$. For Base-to-novel generalization, we instead apply the prior component only to the novel-class slice so that the teacher's probability mass on base classes is preserved. Writing $\mathcal{Y}_b$ and $\mathcal{Y}_n$ for the base and novel classes, we define
\begin{equation}
q_i^P(y)=
\begin{cases}
q_i^T(y), & y\in\mathcal{Y}_b,\\
\displaystyle
\sum_{y'\in\mathcal{Y}_n}q_i^T(y')
\frac{p_i^P(y)}{\sum_{y'\in\mathcal{Y}_n}p_i^P(y')}, & y\in\mathcal{Y}_n.
\end{cases}
\end{equation}
where $q_i^T=\softmax(z_i^T/\tau_i)$ and $p_i^P=\softmax(z_i^P/\tau_i)$. This keeps the adapted teacher in control of base knowledge while drawing on the zero-shot prior where open-vocabulary information is most useful. The novel-class slice relies only on class names through the frozen zero-shot classifier. No novel-class training images or target-domain test samples are ever used to update the student or the policy.

\begin{algorithm}[t]
\caption{\ours{} Training Procedure}
\label{alg:onpokd}
\begin{algorithmic}[1]
\Require Teacher $T$, student $S_{\psi}$, zero-shot prior $P$, policy $f_{\theta}$
\Require Training stream $\mathcal{D}_{\mathrm{tr}}$, validation stream $\mathcal{D}_{\mathrm{val}}$
\Require Base mixture $b$, feasible set $\mathcal{C}$, feedback interval $K$
\Ensure Distilled student $S_{\psi}$
\State Initialize $f_{\theta}$ with a teacher-dominant base action
\For{training step $t=1,2,\ldots$}
    \State Draw a training minibatch $\mathcal{B}_t=\{(x_i,y_i)\}_{i=1}^{B}$
    \State Compute logits and image features from $T$, $S_{\psi}$, and $P$
    \State Build state $s_i$ from reliability and disagreement cues
    \State Obtain raw policy scores $(r_i^\lambda,r_i^w,r_i^\tau)=f_\theta(s_i)$
    \State Estimate need scores $(\eta_i^P,\eta_i^H)$ by Eq.~\eqref{eq:need-scores}
    \State Compute $\lambda_i$, $w_i$, and $\tau_i$ by Eqs.~\eqref{eq:policy-mixture}--\eqref{eq:weight-temp}
    \State Construct adaptive target $q_i$ by Eq.~\eqref{eq:adaptive-target}
    \State Update $\psi$ with $\nabla_{\psi}\mathcal{L}_{\ours}$
    \If{$t \bmod K = 0$}
        \State Draw a validation minibatch $\mathcal{V}_t\subset\mathcal{D}_{\mathrm{val}}$
        \State Evaluate $T$, $S_{\psi}$, and $P$ without updating $\psi$
        \State Build $g_i$ and target action $a_i^\star=F_{\mathrm{val}}(g_i)$
        \State Update $\theta$ with $\nabla_{\theta}\mathcal{L}_{\mathrm{policy}}$
    \EndIf
\EndFor
\end{algorithmic}
\end{algorithm}

The student is trained by
\begin{equation}
    \mathcal{L}_{\ours}
    =
    \frac{1}{B}
    \sum_{i=1}^{B}
    w_i \tau_i^2
    \KL
    \left(
    q_i
    \,\middle\|\,
    \softmax(z_i^S / \tau_i)
    \right),
    \label{eq:onpokd-loss}
\end{equation}
where $B$ is the minibatch size, $w_i$ is a normalized sample weight constrained to a bounded interval, and $\tau_i$ is a bounded sample-wise temperature. To isolate the effect of online target selection, the main configuration can fix $w_i$ and $\tau_i$. An optional feature-level distillation term between teacher and student image features can also be retained. It is orthogonal to the policy and is not part of the target-selection mechanism.

\subsection{Validation Feedback Policy Optimization}

The policy is learned online from a held-out validation stream drawn from the same adaptation source as the student training data. In Base-to-novel generalization this stream contains only base-class examples from the adaptation split, and novel-class test images are never used for policy learning. In Cross-dataset transfer both student training and policy feedback come from the ImageNet source split, with the target datasets reserved for final evaluation. This protocol matters, because the policy should improve transfer behavior without receiving target-domain supervision or optimizing directly on the reported test sets.

At each feedback step, we evaluate the teacher, student, and zero-shot prior on validation samples. We then construct a target action from their correctness and confidence. Let
\begin{equation}
g_i =
\left[
e_T,e_S,e_P,c_T,c_S,c_P,m_T
\right],
\quad
e_M=\indicator[\hat{y}_M=y_i],
\end{equation}
collect validation feedback signals, where $e_M$ is the correctness of model $M\in\{T,S,P\}$. The target action is generated by a deterministic feedback map
\begin{equation}
a_i^\star =
\left(\lambda_i^\star,w_i^\star,\tau_i^\star\right)
=F_{\mathrm{val}}(g_i).
\end{equation}
The feedback map raises the prior coefficient when the prior succeeds and the teacher is unreliable, and raises the hard-label coefficient when both soft sources are unreliable or the student remains uncertain. The sample-weight target is reduced for high-risk samples, and the temperature target stays close to the base distillation temperature unless temperature learning is enabled. The constants in $F_{\mathrm{val}}$ are hyperparameters selected on source validation data. The controller is therefore best understood as a validation-supervised action predictor rather than an unconstrained reinforcement-learning agent.

$\phi_P$, $\phi_H$, and $F_{\mathrm{val}}$ are all deterministic, non-trainable maps. $\phi_P$ takes the teacher's normalized entropy $h_T$, the teacher-prior top-label agreement indicator $d_{TP}^{\argmax}$, and a clipped teacher-prior KL term $d_{TP}^{\KL}$, and returns a clipped monotone mixture over them. $\phi_H$ takes the teacher's normalized entropy $h_T$, the teacher-student top-label agreement indicator $d_{TS}^{\argmax}$, and the student's normalized entropy $h_S$, and returns the same kind of clipped monotone mixture. $F_{\mathrm{val}}$ is a deterministic map from per-sample validation signals $g_i$ to a target action $(\lambda_i^\star, w_i^\star, \tau_i^\star)$, computed as four clipped monotone sub-targets. (i) A teacher-misuse risk score from the teacher's correctness and margin, (ii) a sample-weight target from validation correctness, (iii) a prior-coefficient target from the teacher's gap relative to the prior, and (iv) a hard-label coefficient target from the joint unreliability of the teacher and prior. The mixing weights in $\phi_P$ and $\phi_H$ are positive hyperparameters that sum to one. The auxiliary caps, action bounds, and the rescaling rule for $F_{\mathrm{val}}$ are listed in the released configuration.

The policy loss matches the predicted action to this feedback target.
\begin{equation}
\begin{aligned}
\mathcal{L}_{\mathrm{policy}}
= {}&
\alpha_\lambda \|\lambda_i-\lambda_i^\star\|_2^2
+\alpha_w \|w_i-w_i^\star\|_2^2\\
&+\alpha_\tau \|\tau_i-\tau_i^\star\|_2^2
+\alpha_r \mathcal{R}(\pi_\theta),
\end{aligned}
\end{equation}
where $\mathcal{R}(\pi_\theta)$ regularizes the action toward the conservative base behavior. This validation-feedback objective is decoupled from the training KD loss. It does not back-propagate through that loss or through the student optimizer.

The teacher-dominant initialization is a safety condition for early training. With the final controller layer initialized to zero, the controller's raw contribution is initially neutral and the base simplex action stays teacher-dominant, while the prior and hard-label caps prevent an untrained controller from replacing a useful teacher target with a random auxiliary target. Validation feedback follows a periodic schedule rather than running at every step. At each feedback step the teacher, student, and frozen prior are evaluated on a held-out source-validation minibatch in evaluation mode, and only the controller is updated by the action-matching loss. The schedule reduces minibatch noise and amortizes the cost of the extra evaluation while still allowing the policy to track the changing student later in training.

The default procedure thus separates policy learning from student-gradient updates. On ordinary training batches the controller selects targets for student optimization, whereas on feedback steps validation signals update the controller. This separation prevents the policy from exploiting short-term training-loss reductions and keeps the learned action aligned with validation reliability.

\subsection{Training and Inference}

As a training-time target-construction mechanism, \ours{} imposes no specific teacher or student architecture. The teacher can be any adapted vision-language model that exposes class logits and image features, and the prior is obtained from the frozen zero-shot vision-language model using class-name prompts. During training, the policy controller reads reliability and disagreement cues from the teacher, student, and prior and uses them to construct the adaptive target in Eq.~\eqref{eq:adaptive-target}. At inference, the controller and the zero-shot prior branch are removed, and the distilled student produces the final prediction with the same adapted text classifier as the underlying teacher-student VLM adaptation pipeline. In this way, \ours{} improves the supervision signal during distillation while adding no policy evaluation, prior fusion, or extra model capacity at test time.

\begin{table*}[!h]
    \centering
    \captionsetup{skip=5pt}
    \caption{Base-to-novel generalization on the standard eleven-dataset \clip{} benchmark. Each block reports Base, Novel, and HM accuracy, and $\Delta$ shows the gain over PromptKD.}
    \renewcommand{\arraystretch}{1}
    \vspace{5pt}

    \begin{minipage}{\textwidth}
        \centering
        \begin{minipage}[t]{0.32\linewidth}
            \centering
            \vspace{0pt}
            \footnotesize(a) Average over 11 datasets
            \vspace{2pt}

            \resizebox{\linewidth}{!}{
            \begin{tabular}{cccc}
                \hline
                ViT-B/16 & Base  & Novel & HM \\
                \hline
                CLIP & 69.34 & 74.22 & 71.70 \\
                CoOp & 82.69 & 63.22 & 71.66 \\
                CoCoOp & 80.47 & 72.30 & 76.17 \\
                MaPLe & 82.28 & 75.14 & 78.55 \\
                PromptSRC & 84.26 & 76.10 & 79.97 \\
                PromptKD & 86.96 & 80.73 & 83.73 \\
                \hline
                \cellcolor{lightgray!30}\ours{} & \cellcolor{lightgray!30}87.26 & \cellcolor{lightgray!30}82.13 & \cellcolor{lightgray!30}84.62 \\
                $\Delta$ & \textcolor{customred}{\textbf{+0.30}} & \textcolor{customred}{\textbf{+1.40}} & \textcolor{customred}{\textbf{+0.89}} \\
                \hline
            \end{tabular}
            }
        \end{minipage}
        \hfill
        \begin{minipage}[t]{0.32\linewidth}
            \centering
            \vspace{0pt}
            \footnotesize(b) ImageNet
            \vspace{2pt}

            \resizebox{\linewidth}{!}{
            \begin{tabular}{cccc}
                \hline
                ViT-B/16 & Base  & Novel & HM \\
                \hline
                CLIP & 72.43 & 68.14 & 70.22 \\
                CoOp & 76.47 & 67.88 & 71.92 \\
                CoCoOp & 75.98 & 70.43 & 73.10 \\
                MaPLe & 76.66 & 70.54 & 73.47 \\
                PromptSRC & 77.60 & 70.73 & 74.01 \\
                PromptKD & 80.83 & 74.66 & 77.62 \\
                \hline
                \cellcolor{lightgray!30}\ours{} & \cellcolor{lightgray!30}80.90 & \cellcolor{lightgray!30}75.20 & \cellcolor{lightgray!30}77.95 \\
                $\Delta$ & \textcolor{customred}{\textbf{+0.07}} & \textcolor{customred}{\textbf{+0.54}} & \textcolor{customred}{\textbf{+0.33}} \\
                \hline
            \end{tabular}
            }
        \end{minipage}
        \hfill
        \begin{minipage}[t]{0.32\linewidth}
            \centering
            \vspace{0pt}
            \footnotesize(c) Caltech101
            \vspace{2pt}

            \resizebox{\linewidth}{!}{
            \begin{tabular}{cccc}
                \hline
                ViT-B/16 & Base  & Novel & HM \\
                \hline
                CLIP & 96.84 & 94.00 & 95.40 \\
                CoOp & 98.00 & 89.81 & 93.73 \\
                CoCoOp & 97.96 & 93.81 & 95.84 \\
                MaPLe & 97.74 & 94.36 & 96.02 \\
                PromptSRC & 98.10 & 94.03 & 96.02 \\
                PromptKD & 98.91 & 96.65 & 97.77 \\
                \hline
                \cellcolor{lightgray!30}\ours{} & \cellcolor{lightgray!30}99.12 & \cellcolor{lightgray!30}97.13 & \cellcolor{lightgray!30}98.11 \\
                $\Delta$ & \textcolor{customred}{\textbf{+0.21}} & \textcolor{customred}{\textbf{+0.48}} & \textcolor{customred}{\textbf{+0.34}} \\
                \hline
            \end{tabular}
            }
        \end{minipage}
    \end{minipage}
    \vspace{8pt}

    \begin{minipage}{\textwidth}
        \centering
        \begin{minipage}[t]{0.32\linewidth}
            \centering
            \vspace{0pt}
            \footnotesize(d) OxfordPets
            \vspace{2pt}

            \resizebox{\linewidth}{!}{
            \begin{tabular}{cccc}
                \hline
                ViT-B/16 & Base  & Novel & HM \\
                \hline
                CLIP & 91.17 & 97.26 & 94.12 \\
                CoOp & 93.67 & 95.29 & 94.47 \\
                CoCoOp & 95.20 & 97.69 & 96.43 \\
                MaPLe & 95.43 & 97.76 & 96.58 \\
                PromptSRC & 95.33 & 97.30 & 96.30 \\
                PromptKD & 96.30 & 98.01 & 97.15 \\
                \hline
                \cellcolor{lightgray!30}\ours{} & \cellcolor{lightgray!30}96.34 & \cellcolor{lightgray!30}98.31 & \cellcolor{lightgray!30}97.32 \\
                $\Delta$ & \textcolor{customred}{\textbf{+0.04}} & \textcolor{customred}{\textbf{+0.30}} & \textcolor{customred}{\textbf{+0.17}} \\
                \hline
            \end{tabular}
            }
        \end{minipage}
        \hfill
        \begin{minipage}[t]{0.32\linewidth}
            \centering
            \vspace{0pt}
            \footnotesize(e) StanfordCars
            \vspace{2pt}

            \resizebox{\linewidth}{!}{
            \begin{tabular}{cccc}
                \hline
                ViT-B/16 & Base  & Novel & HM \\
                \hline
                CLIP & 63.37 & 74.89 & 68.65 \\
                CoOp & 78.12 & 60.40 & 68.13 \\
                CoCoOp & 70.49 & 73.59 & 72.01 \\
                MaPLe & 72.94 & 74.00 & 73.47 \\
                PromptSRC & 78.27 & 74.97 & 76.58 \\
                PromptKD & 82.80 & 83.37 & 83.08 \\
                \hline
                \cellcolor{lightgray!30}\ours{} & \cellcolor{lightgray!30}83.19 & \cellcolor{lightgray!30}84.64 & \cellcolor{lightgray!30}83.91 \\
                $\Delta$ & \textcolor{customred}{\textbf{+0.39}} & \textcolor{customred}{\textbf{+1.27}} & \textcolor{customred}{\textbf{+0.83}} \\
                \hline
            \end{tabular}
            }
        \end{minipage}
        \hfill
        \begin{minipage}[t]{0.32\linewidth}
            \centering
            \vspace{0pt}
            \footnotesize(f) Flowers102
            \vspace{2pt}

            \resizebox{\linewidth}{!}{
            \begin{tabular}{cccc}
                \hline
                ViT-B/16 & Base  & Novel & HM \\
                \hline
                CLIP & 72.08 & 77.80 & 74.83 \\
                CoOp & 97.60 & 59.67 & 74.06 \\
                CoCoOp & 94.87 & 71.75 & 81.71 \\
                MaPLe & 95.92 & 72.46 & 82.56 \\
                PromptSRC & 98.07 & 76.50 & 85.95 \\
                PromptKD & 99.42 & 82.62 & 90.24 \\
                \hline
                \cellcolor{lightgray!30}\ours{} & \cellcolor{lightgray!30}99.33 & \cellcolor{lightgray!30}83.28 & \cellcolor{lightgray!30}90.60 \\
                $\Delta$ & -0.09 & \textcolor{customred}{\textbf{+0.66}} & \textcolor{customred}{\textbf{+0.36}} \\
                \hline
            \end{tabular}
            }
        \end{minipage}
    \end{minipage}
    \vspace{8pt}

    \begin{minipage}{\textwidth}
        \centering
        \begin{minipage}[t]{0.32\linewidth}
            \centering
            \vspace{0pt}
            \footnotesize(g) Food101
            \vspace{2pt}

            \resizebox{\linewidth}{!}{
            \begin{tabular}{cccc}
                \hline
                ViT-B/16 & Base  & Novel & HM \\
                \hline
                CLIP & 90.10 & 91.22 & 90.66 \\
                CoOp & 88.33 & 82.26 & 85.19 \\
                CoCoOp & 90.70 & 91.29 & 90.99 \\
                MaPLe & 90.71 & 92.05 & 91.38 \\
                PromptSRC & 90.67 & 91.53 & 91.10 \\
                PromptKD & 92.43 & 93.68 & 93.05 \\
                \hline
                \cellcolor{lightgray!30}\ours{} & \cellcolor{lightgray!30}92.62 & \cellcolor{lightgray!30}93.94 & \cellcolor{lightgray!30}93.28 \\
                $\Delta$ & \textcolor{customred}{\textbf{+0.19}} & \textcolor{customred}{\textbf{+0.26}} & \textcolor{customred}{\textbf{+0.23}} \\
                \hline
            \end{tabular}
            }
        \end{minipage}
        \hfill
        \begin{minipage}[t]{0.32\linewidth}
            \centering
            \vspace{0pt}
            \footnotesize(h) FGVCAircraft
            \vspace{2pt}

            \resizebox{\linewidth}{!}{
            \begin{tabular}{cccc}
                \hline
                ViT-B/16 & Base  & Novel & HM \\
                \hline
                CLIP & 27.19 & 36.29 & 31.09 \\
                CoOp & 40.44 & 22.30 & 28.75 \\
                CoCoOp & 33.41 & 23.71 & 27.74 \\
                MaPLe & 37.44 & 35.61 & 36.50 \\
                PromptSRC & 42.73 & 37.87 & 40.15 \\
                PromptKD & 49.12 & 41.81 & 45.17 \\
                \hline
                \cellcolor{lightgray!30}\ours{} & \cellcolor{lightgray!30}50.43 & \cellcolor{lightgray!30}45.17 & \cellcolor{lightgray!30}47.66 \\
                $\Delta$ & \textcolor{customred}{\textbf{+1.31}} & \textcolor{customred}{\textbf{+3.36}} & \textcolor{customred}{\textbf{+2.49}} \\
                \hline
            \end{tabular}
            }
        \end{minipage}
        \hfill
        \begin{minipage}[t]{0.32\linewidth}
            \centering
            \vspace{0pt}
            \footnotesize(i) SUN397
            \vspace{2pt}

            \resizebox{\linewidth}{!}{
            \begin{tabular}{cccc}
                \hline
                ViT-B/16 & Base  & Novel & HM \\
                \hline
                CLIP & 69.36 & 75.35 & 72.23 \\
                CoOp & 80.60 & 65.89 & 72.51 \\
                CoCoOp & 79.74 & 76.86 & 78.27 \\
                MaPLe & 80.82 & 78.70 & 79.75 \\
                PromptSRC & 82.67 & 78.47 & 80.52 \\
                PromptKD & 83.69 & 81.54 & 82.60 \\
                \hline
                \cellcolor{lightgray!30}\ours{} & \cellcolor{lightgray!30}83.92 & \cellcolor{lightgray!30}82.33 & \cellcolor{lightgray!30}83.12 \\
                $\Delta$ & \textcolor{customred}{\textbf{+0.23}} & \textcolor{customred}{\textbf{+0.79}} & \textcolor{customred}{\textbf{+0.52}} \\
                \hline
            \end{tabular}
            }
        \end{minipage}
    \end{minipage}
    \vspace{8pt}

    \begin{minipage}{\textwidth}
        \centering
        \begin{minipage}[t]{0.32\linewidth}
            \centering
            \vspace{0pt}
            \footnotesize(j) DTD
            \vspace{2pt}

            \resizebox{\linewidth}{!}{
            \begin{tabular}{cccc}
                \hline
                ViT-B/16 & Base  & Novel & HM \\
                \hline
                CLIP & 53.24 & 59.90 & 56.37 \\
                CoOp & 79.44 & 41.18 & 54.24 \\
                CoCoOp & 77.01 & 56.00 & 64.85 \\
                MaPLe & 80.36 & 59.18 & 68.16 \\
                PromptSRC & 83.37 & 62.97 & 71.75 \\
                PromptKD & 85.84 & 71.37 & 77.94 \\
                \hline
                \cellcolor{lightgray!30}\ours{} & \cellcolor{lightgray!30}86.64 & \cellcolor{lightgray!30}73.59 & \cellcolor{lightgray!30}79.58 \\
                $\Delta$ & \textcolor{customred}{\textbf{+0.80}} & \textcolor{customred}{\textbf{+2.22}} & \textcolor{customred}{\textbf{+1.64}} \\
                \hline
            \end{tabular}
            }
        \end{minipage}
        \hfill
        \begin{minipage}[t]{0.32\linewidth}
            \centering
            \vspace{0pt}
            \footnotesize(k) EuroSAT
            \vspace{2pt}

            \resizebox{\linewidth}{!}{
            \begin{tabular}{cccc}
                \hline
                ViT-B/16 & Base  & Novel & HM \\
                \hline
                CLIP & 56.48 & 64.05 & 60.03 \\
                CoOp & 92.19 & 54.74 & 68.69 \\
                CoCoOp & 87.49 & 60.04 & 71.21 \\
                MaPLe & 94.07 & 73.23 & 82.35 \\
                PromptSRC & 92.90 & 73.90 & 82.32 \\
                PromptKD & 97.54 & 82.08 & 89.14 \\
                \hline
                \cellcolor{lightgray!30}\ours{} & \cellcolor{lightgray!30}97.64 & \cellcolor{lightgray!30}87.01 & \cellcolor{lightgray!30}92.02 \\
                $\Delta$ & \textcolor{customred}{\textbf{+0.10}} & \textcolor{customred}{\textbf{+4.93}} & \textcolor{customred}{\textbf{+2.88}} \\
                \hline
            \end{tabular}
            }
        \end{minipage}
        \hfill
        \begin{minipage}[t]{0.32\linewidth}
            \centering
            \vspace{0pt}
            \footnotesize(l) UCF101
            \vspace{2pt}

            \resizebox{\linewidth}{!}{
            \begin{tabular}{cccc}
                \hline
                ViT-B/16 & Base  & Novel & HM \\
                \hline
                CLIP & 70.53 & 77.50 & 73.85 \\
                CoOp & 84.69 & 56.05 & 67.46 \\
                CoCoOp & 82.33 & 73.45 & 77.64 \\
                MaPLe & 83.00 & 78.66 & 80.77 \\
                PromptSRC & 87.10 & 78.80 & 82.74 \\
                PromptKD & 89.71 & 82.27 & 85.83 \\
                \hline
                \cellcolor{lightgray!30}\ours{} & \cellcolor{lightgray!30}89.78 & \cellcolor{lightgray!30}82.84 & \cellcolor{lightgray!30}86.17 \\
                $\Delta$ & \textcolor{customred}{\textbf{+0.07}} & \textcolor{customred}{\textbf{+0.57}} & \textcolor{customred}{\textbf{+0.34}} \\
                \hline
            \end{tabular}
            }
        \end{minipage}
    \end{minipage}
    \label{table:onpokd_base2novel}
    \vspace{-10pt}
\end{table*}

\section{Experiments}

\subsection{Experimental Setup}

\paragraph{Base-to-Novel Generalization.}
The Base-to-novel generalization setting tests whether adaptation preserves open-vocabulary generalization~\citep{zhou2022learning,zhou2022conditional,khattak2023maple}. Each dataset is split into base and novel classes. The model is adapted only with base-class training data and then evaluated on both base and novel test classes. This setting is well aligned with our goal, as it directly exposes the trade-off between teacher-driven adaptation and preservation of the original zero-shot prior. Following common practice, we report base accuracy, novel accuracy, and their harmonic mean (HM), which summarizes the balance between adapted-class discrimination and novel-class transfer.

\paragraph{Cross-Dataset Evaluation.}
The Cross-dataset evaluation setting tests robustness under dataset shift~\citep{recht2019imagenet,hendrycks2021many}. The model is adapted on ImageNet and then transferred to other recognition datasets without using their training images, which probes whether the distilled student learns a transferable decision rule or instead overfits to ImageNet classes and visual statistics. We report top-1 accuracy on each target dataset and the average accuracy across all target datasets.

\paragraph{Datasets.}
For Base-to-novel generalization, we follow the standard eleven-dataset benchmark used in vision-language prompt learning. The benchmark includes ImageNet~\citep{deng2009imagenet}, Caltech101~\citep{fei2004learning}, OxfordPets~\citep{parkhi2012cats}, StanfordCars~\citep{krause20133d}, Flowers102~\citep{nilsback2008automated}, Food101~\citep{bossard2014food}, FGVCAircraft~\citep{maji2013fine}, SUN397~\citep{xiao2010sun}, DTD~\citep{cimpoi2014describing}, EuroSAT~\citep{helber2019eurosat}, and UCF101~\citep{soomro2012ucf101}. These datasets span objects, fine-grained categories, textures, scenes, actions, and remote-sensing images. For Cross-dataset evaluation, ImageNet serves as the source dataset and the adapted model is evaluated on the other ten. This diversity lets us assess both class-shift and domain-shift transfer.

\paragraph{Implementation Details.}
All comparisons use the ViT-B/16 \clip{} backbone~\citep{radford2021learning}. We compare \ours{} against zero-shot \clip{}, the prompt-learning methods CoOp, CoCoOp, MaPLe, and PromptSRC~\citep{zhou2022learning,zhou2022conditional,khattak2023maple,khattak2023self}, and the distillation baseline PromptKD~\citep{li2024promptkd}. For a controlled comparison, \ours{} leaves the teacher, the student, and the test-time inference architecture unchanged. The controller is a two-layer multilayer perceptron with layer normalization whose final layer is initialized to zero, so its initial behavior is the conservative teacher-dominant action in Algorithm~\ref{alg:onpokd}. Hyperparameters are selected with source or base validation feedback and then kept fixed for all reported evaluations. Because the only difference from the underlying distillation pipeline is the on-policy for target construction, the improvements in the following tables reflect adaptive target construction rather than extra model capacity, target-domain tuning, or additional inference-time modules. The controller architecture and base action are shared across the two evaluation settings, while the policy hyperparameters take two pre-specified configurations that are fixed before any target-dataset evaluation. The two settings differ because the Base-to-novel and Cross-dataset tasks expose different shifts, not because the controller is tuned to individual target datasets. The training time and memory overhead on the two representative datasets used in the ablation study are reported in Table~\ref{tab:efficiency}.

\begin{table}[t]
\centering
\textcolor{customred}{\caption{Training time and memory overhead of \ours{} relative to PromptKD on the two representative datasets used throughout the ablation studies.}\label{tab:efficiency}}
\scriptsize

\begin{tabular}{lccc}
\toprule
\textbf{Dataset} & \textbf{HM gain} & \textbf{Training time} & \textbf{Memory} \\
\midrule
FGVCAircraft & $+2.49$ & $+25\%$ & $+51\%$ \\
EuroSAT       & $+2.88$ & $+19\%$ & $+51\%$ \\
\bottomrule
\end{tabular}
\end{table}

\subsection{Base-to-Novel Generalization}

Table~\ref{table:onpokd_base2novel} reports Base-to-novel generalization results across eleven datasets. On average, \ours{} improves the HM over PromptKD from 83.73 to 84.62, with a larger gain on novel classes (+1.40) than on base classes (+0.30). This pattern matches the motivation of \ours{}. The policy keeps teacher-centered supervision for reliable base-class knowledge while injecting zero-shot prior and hard-label guidance when the adapted teacher is less reliable for transfer. The gains are broadly distributed rather than concentrated on one dataset. \ours{} improves HM on all eleven benchmarks, and the improvement is especially clear on FGVCAircraft (+2.49 HM), DTD (+1.64 HM), EuroSAT (+2.88 HM), and StanfordCars (+0.83 HM), while even ImageNet rises by +0.33 HM. The strongest gains appear precisely on datasets where the Base-to-novel generalization gap is hardest to close, which suggests that sample-wise policy control better balances adapted teacher knowledge against zero-shot prior knowledge.

\subsection{Cross-Dataset Transfer}

\begin{table*}[t!]
\centering
\caption{Cross-dataset accuracy (\%) on ten target datasets. PromptKD and OnPoKD distill ImageNet-adapted teachers using unlabeled target-domain training images, while other methods transfer directly from ImageNet. Avg. denotes the mean target accuracy.}
\scriptsize
\renewcommand{\arraystretch}{1}
\setlength{\tabcolsep}{3pt}
\resizebox{\textwidth}{!}{
\begin{tabular}{lccccccccccc}
\toprule
\textbf{Method} & \textbf{Avg.} & \textbf{Caltech101} & \textbf{OxfordPets} & \textbf{Cars} & \textbf{Flowers102} & \textbf{Food101} & \textbf{Aircraft} & \textbf{SUN397} & \textbf{DTD} & \textbf{EuroSAT} & \textbf{UCF101} \\
\midrule
CoOp & 63.88 & 93.70 & 89.14 & 64.51 & 68.71 & 85.30 & 18.47 & 64.15 & 41.92 & 46.39 & 66.55 \\
CoCoOp & 65.84 & 94.43 & 90.14 & 65.32 & 71.88 & 86.06 & 22.94 & 67.36 & 45.73 & 46.37 & 68.21 \\
MaPLe & 66.30 & 93.53 & 90.49 & 65.57 & 72.23 & 86.20 & 24.74 & 67.01 & 46.49 & 48.06 & 68.69 \\
PromptSRC & 65.81 & 93.60 & 90.25 & 65.70 & 70.25 & 86.15 & 23.90 & 67.10 & 46.87 & 45.50 & 68.75 \\
PromptKD & 71.33 & \textbf{93.61} & 91.59 & 73.93 & 75.33 & \textbf{88.84} & 26.24 & 68.57 & 55.08 & 63.74 & \textbf{76.39} \\
\midrule
\ours{} & \textbf{72.66} & 93.04 & \textbf{92.21} & \textbf{76.04} & \textbf{76.87} & 88.52 & \textbf{28.03} & \textbf{68.92} & \textbf{58.19} & \textbf{69.43} & 75.32 \\
$\Delta$ & \textcolor{customred}{\textbf{+1.33}} & -0.57 & \textcolor{customred}{\textbf{+0.62}} & \textcolor{customred}{\textbf{+2.11}} & \textcolor{customred}{\textbf{+1.54}} & -0.32 & \textcolor{customred}{\textbf{+1.79}} & \textcolor{customred}{\textbf{+0.35}} & \textcolor{customred}{\textbf{+3.11}} & \textcolor{customred}{\textbf{+5.69}} & -1.07 \\
\bottomrule
\end{tabular}
}
\label{tab:cross_dataset}
\end{table*}

Table~\ref{tab:cross_dataset} evaluates Cross-dataset transfer from ImageNet to ten target datasets. This benchmark is stricter than Base-to-novel generalization, since the target datasets are never seen during adaptation. \ours{} reaches the best average accuracy of 72.66, improving PromptKD by 1.33\%. The policy therefore does not merely overfit the source-domain adaptation objective. It strengthens the distilled student's ability to transfer across datasets. \ours{} obtains the strongest accuracy on seven of the ten target datasets, with the largest gains on EuroSAT (+5.69), DTD (+3.11), Cars (+2.11), Aircraft (+1.79), and Flowers102 (+1.54), all of which differ from ImageNet in visual domain or fine-grained label semantics, where a fixed source-trained teacher signal is brittle. The small decreases on Caltech101 (-0.57), Food101 (-0.32), and UCF101 (-1.07) occur on datasets where PromptKD is already strong, a trade-off suggesting that future policy variants may need more conservative intervention when the teacher and prior are both confident.

\begin{table*}[t]
\centering
\caption{Controlled removal study for the target-construction policy. T-KD uses only the adapted teacher, Fixed Mix removes sample-wise decisions, and the remaining variants disable one policy source or update signal.}
\scriptsize
\setlength{\tabcolsep}{10pt}
\resizebox{0.8\textwidth}{!}{
\begin{tabular}{lcccccccc}
\toprule
\multirow{3}{*}{\textbf{Variant}} &
\multicolumn{6}{c}{\textbf{Base-to-Novel}} &
\multicolumn{2}{c}{\textbf{Cross-Dataset}} \\
\cmidrule(lr){2-7}\cmidrule(lr){8-9}
& \multicolumn{3}{c}{\textbf{FGVCAircraft}} &
\multicolumn{3}{c}{\textbf{EuroSAT}} &
\textbf{DTD} & \textbf{EuroSAT} \\
\cmidrule(lr){2-4}\cmidrule(lr){5-7}\cmidrule(lr){8-9}
& \textbf{Base} & \textbf{Novel} & \textbf{HM}
& \textbf{Base} & \textbf{Novel} & \textbf{HM}
& \textbf{Acc.} & \textbf{Acc.} \\
\midrule
T-KD & 47.71 & 40.84 & 44.01 & 97.13 & 85.24 & 90.80 & 53.19 & 62.55 \\
Fixed Mix & 48.12 & 40.47 & 43.96 & 96.46 & 84.25 & 89.94 & 53.00 & 64.12 \\
w/o Prior & 48.90 & 43.71 & 46.16 & 96.98 & 81.80 & 88.75 & 57.32 & 69.97 \\
w/o Label & 47.45 & 41.03 & 44.01 & 96.58 & 80.36 & 87.73 & 53.12 & 66.96 \\
w/o Val. & 18.81 & 28.73 & 22.74 & 96.51 & 84.78 & 90.27 & 53.29 & 65.35 \\
w/o Prog. & 48.90 & 43.90 & 46.27 & 97.52 & 85.64 & 91.19 & 57.11 & \textbf{70.45} \\
Full & \textbf{50.43} & \textbf{45.17} & \textbf{47.66} & \textbf{97.64} & \textbf{87.01} & \textbf{92.02} & \textbf{58.19} & 69.43 \\
\bottomrule
\end{tabular}
}
\label{tab:ablation_components}
\end{table*}

\subsection{Ablation Study}

We conduct ablations from three complementary perspectives (the supervision components, the policy action space, and the controller state cues) using representative datasets from both evaluation settings. For Base-to-novel generalization we use FGVCAircraft and EuroSAT, which emphasize fine-grained class transfer and domain-shifted recognition. For Cross-dataset transfer we use DTD and EuroSAT, whose target domains differ substantially from the ImageNet source. This design tests whether the policy-controlled target is useful under both class shift and dataset shift.

\subsubsection{Contribution of Supervision Sources}

We first examine whether the gain comes from the extra supervision sources or from the on-policy that controls them. This is an important distinction, since simply adding the zero-shot prior or hard labels could improve robustness even without sample-wise adaptation. Under the same teacher-student setup, Table~\ref{tab:ablation_components} compares four groups (teacher-only distillation, a fixed teacher-prior-label mixture, source-removal variants, and policy-learning variants). T-KD reduces \ours{} to ordinary teacher-only distillation, using the adapted teacher as the only soft target. Fixed Mix keeps the same three supervision sources as \ours{} but combines them with a fixed teacher-prior-label mixture, thereby removing sample-wise policy adaptation. The w/o Prior and w/o Label variants drop the zero-shot prior and hard-label components to test whether the gains come from open-vocabulary prior knowledge or from explicit label anchoring. Finally, the w/o Val. and w/o Prog. variants retain the target sources but remove validation-feedback updates or progress-aware modulation, testing whether online learning is needed for reliable target construction. Table~\ref{tab:ablation_components} shows that the full design gives the best overall balance. Relative to T-KD, \ours{} improves FGVCAircraft HM by 3.65\%, EuroSAT HM by 1.22\%, DTD accuracy by 5.00\%, and EuroSAT Cross-dataset accuracy by 6.88\%, gains that cannot be attributed to teacher imitation alone. Fixed Mix remains consistently weaker than the adaptive variants, which confirms that the supervision sources alone are not enough. The policy must decide when to trust each one. The source-removal variants reveal distinct roles for the prior and the label. Removing labels causes the largest drop on Cross-dataset transfer, especially on DTD and EuroSAT, indicating that hard labels anchor the controller when teacher and prior disagree. Removing the prior instead hurts FGVCAircraft and EuroSAT Base-to-novel generalization more clearly, indicating that the zero-shot prior mainly protects open-vocabulary transfer. The w/o Val. variant collapses FGVCAircraft HM, confirming that validation feedback is needed to avoid unstable decisions on fine-grained classes. Specifically, removing validation feedback reduces FGVCAircraft HM from 47.66 to 22.74 (a reduction of 24.92\%) and EuroSAT HM from 92.02 to 90.27 (a reduction of 1.75\%). The gap reflects the dataset-dependent role of held-out feedback. On fine-grained classes with narrow inter-class margins and sample-specific teacher mistakes, confidence-based reliability cues alone become insufficient, so direct correctness signals from a held-out source-validation stream are necessary for the controller to suppress an unsafe intervention. On EuroSAT, by contrast, state-based uncertainty and disagreement signals remain informative because the class cues are stronger and more coherent, so the marginal value of validation feedback is smaller but still positive. Removing progress-aware modulation has a milder effect: it slightly helps EuroSAT Cross-dataset accuracy yet still lowers FGVCAircraft HM and DTD accuracy. Overall, each source plays a distinct part. The teacher supports task adaptation, the prior supports robustness, the label anchors learning, and feedback keeps the mixture reliable.

\subsubsection{Contribution of Each Policy Action}

\begin{table*}[t]
\centering
\caption{Effect of enabling policy actions one at a time. The comparison isolates target mixing, sample weighting, and temperature control on FGVCAircraft and EuroSAT.}
\scriptsize
\setlength{\tabcolsep}{15pt}
\resizebox{0.85\textwidth}{!}{
\begin{tabular}{lcccccc}
\toprule
\multirow{2}{*}{\textbf{Variant}} &
\multicolumn{3}{c}{\textbf{FGVCAircraft}} &
\multicolumn{3}{c}{\textbf{EuroSAT}} \\
\cmidrule(lr){2-4}\cmidrule(lr){5-7}
& \textbf{Base} & \textbf{Novel} & \textbf{HM}
& \textbf{Base} & \textbf{Novel} & \textbf{HM} \\
\midrule
Fixed KD target & 49.28 & 44.27 & 46.64 & 97.24 & 84.28 & 90.30 \\
Adaptive mix only & 23.59 & 28.43 & 25.78 & 97.52 & 86.69 & 91.79 \\
Adaptive mix + weight & 48.14 & 40.43 & 43.95 & 96.40 & 84.21 & 89.89 \\
Adaptive mix + temp. & 43.64 & 40.13 & 41.81 & 97.02 & 83.23 & 89.60 \\
Full action & \textbf{50.43} & \textbf{45.17} & \textbf{47.66} & \textbf{97.64} & \textbf{87.01} & \textbf{92.02} \\
\bottomrule
\end{tabular}
}
\label{tab:ablation_action_space}
\end{table*}

\begin{table*}[t]
\centering
\caption{Reliability cues used by the policy controller. Each row removes one cue group while keeping the full action space fixed.}
\scriptsize
\setlength{\tabcolsep}{15pt}
\resizebox{0.85\textwidth}{!}{
\begin{tabular}{lcccccc}
\toprule
\multirow{2}{*}{\textbf{Variant}} &
\multicolumn{3}{c}{\textbf{FGVCAircraft}} &
\multicolumn{3}{c}{\textbf{EuroSAT}} \\
\cmidrule(lr){2-4}\cmidrule(lr){5-7}
& \textbf{Base} & \textbf{Novel} & \textbf{HM}
& \textbf{Base} & \textbf{Novel} & \textbf{HM} \\
\midrule
w/o uncertainty cues & 36.73 & 39.29 & 37.97 & 97.26 & 81.56 & 88.72 \\
w/o T-S disagreement & 47.12 & 43.79 & 45.39 & 97.55 & 84.64 & 90.64 \\
w/o T-P conflict & 48.44 & 44.57 & 46.42 & 96.50 & 86.23 & 91.08 \\
w/o feature alignment & 47.78 & 44.33 & 45.99 & 97.12 & 85.05 & 90.69 \\
Full state & \textbf{50.43} & \textbf{45.17} & \textbf{47.66} & \textbf{97.64} & \textbf{87.01} & \textbf{92.02} \\
\bottomrule
\end{tabular}
}
\label{tab:ablation_state_cues}
\end{table*}

\begin{figure*}[t]
    \centering
    \includegraphics[width=0.95\textwidth]{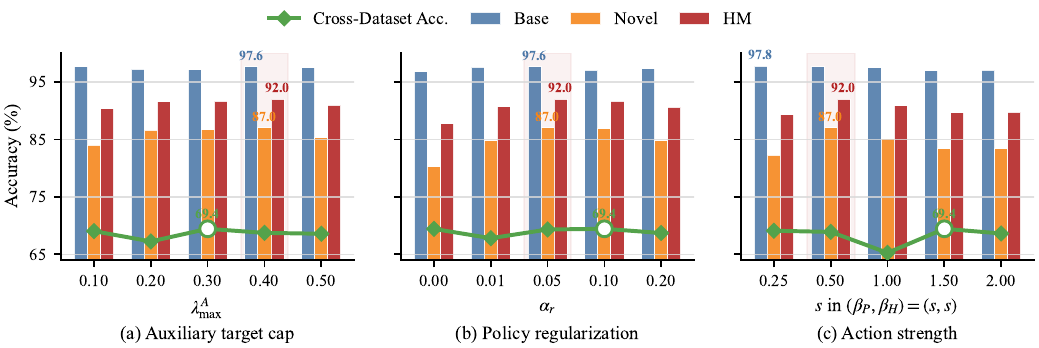}
    \caption{EuroSAT sensitivity sweeps for the main policy controls. Bars show Base-to-novel generalization metrics, and the line shows Cross-dataset accuracy.}
    \label{fig:hyper_sensitivity}
\end{figure*}

\begin{figure}[pos=h,width=\columnwidth]
\centering
\textcolor{customred}{\caption{Learned state-to-action mapping on FGVC-Aircraft (weak teacher) and EuroSAT (strong teacher). The policy increases prior mass with uncertainty on EuroSAT but keeps the prior dormant on FGVC-Aircraft, with all cues producing the expected action directions.}\label{fig:policy_state_action}}
\includegraphics[width=\columnwidth]{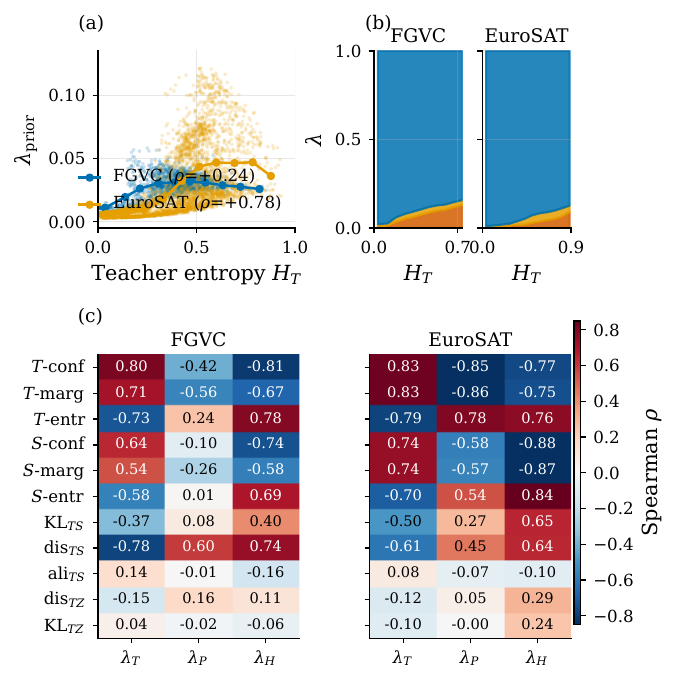}
\end{figure}

We next study which actions the controller needs for effective target construction. The full policy controls three quantities (the mixture over teacher, prior, and label targets, the sample weight, and the distillation temperature). Starting from a fixed KD target, Table~\ref{tab:ablation_action_space} enables these action dimensions one at a time. The ``Adaptive mix only'' variant lets the controller select the teacher-prior-label mixture while fixing the sample weight and temperature. The next two variants add either sample weighting or temperature adaptation. Table~\ref{tab:ablation_action_space} shows that these action dimensions are neither interchangeable nor uniformly useful across tasks. On EuroSAT, whose class-level visual cues are relatively coherent, adaptive mixing already improves HM from 90.30 to 91.79, whereas adding sample weighting or temperature adaptation on top of the mixture yields HMs of 89.89 and 89.60, respectively, suggesting that extra per-sample corrections can disturb an already adequate mixture. FGVCAircraft is more fine-grained with narrower inter-class margins, so adaptive mixing alone drops HM from 46.64 to 25.78. Sample weighting and temperature adaptation recover HM to 43.95 and 41.81 by down-weighting unreliable examples and adjusting the softness of the distillation signal, respectively. The full action space performs best precisely because the three actions play complementary roles. Mixture decides what to trust, weighting decides how much to trust a sample, and temperature decides how sharply to match the target, a coupling that a fixed KD target cannot express.

\subsubsection{Importance of Reliability Cues}

Finally, we analyze which reliability cues the controller needs. \ours{} builds its policy state from four cue groups (uncertainty statistics, teacher-student disagreement, teacher-prior conflict, and feature alignment). Table~\ref{tab:ablation_state_cues} removes each group in turn while keeping the action space fixed. Each ablation isolates one role. Removing uncertainty cues tests whether confidence, margin, and entropy are needed to identify unreliable predictions, removing teacher-student disagreement tests whether the policy must know when the student is misaligned with the adapted teacher, removing teacher-prior conflict tests the role of disagreement between adapted and zero-shot knowledge, and removing feature alignment tests whether representation-level transfer quality adds complementary information. Table~\ref{tab:ablation_state_cues} shows that uncertainty cues are the most important state signal. Removing them drops FGVCAircraft HM by 9.69\% and EuroSAT HM by 3.30\%, far more than any single relational-cue ablation. The controller must first know whether a sample is unreliable, and the remaining cue groups then explain why. Teacher-student disagreement shows whether the student has caught up with the teacher, teacher-prior conflict reveals when adapted and zero-shot knowledge disagree, and feature alignment provides a representation-level check on transfer quality. Their individual drops are smaller but consistent across both datasets, indicating that the best policy state combines confidence with relational cues. A single confidence score is not enough. Figure~\ref{fig:policy_state_action} visualizes the learned state-to-action mapping on the two representative ablation datasets, complementing the row-removal ablations above with a continuous view of how the controller actually responds to its inputs. On EuroSAT, every cue moves its expected action direction. Teacher entropy and teacher-prior disagreement couple most strongly with the prior coefficient, while teacher-student disagreement and student entropy drive the hard-label coefficient. On FGVCAircraft the same cues are largely inert. The prior channel stays nearly dormant across the cue range, and only teacher entropy shows a weak positive slope on the prior coefficient. This pattern matches the row-removal result. Uncertainty cues carry most of the signal on EuroSAT, while on FGVCAircraft the cues themselves are less informative so the controller has less to work with. The figure therefore confirms that the learned policy is dataset-aware rather than a fixed rule, which is consistent with the dataset-dependent gains in Table~\ref{tab:ablation_state_cues}.

\subsection{Hyperparameter Analysis}

We further analyze the sensitivity of \ours{} to key hyperparameters. Because the controller is designed to make bounded interventions during training, the method should not rely on a narrow hyperparameter setting. Figure~\ref{fig:hyper_sensitivity} summarizes sweeps on EuroSAT, a representative domain-shifted benchmark, reporting both Base-to-novel generalization metrics and Cross-dataset transfer accuracy. Two patterns emerge. The auxiliary cap and policy regularization most clearly affect the Base-to-novel generalization trade-off, whereas Cross-dataset transfer is less monotonic and stays sensitive to the exact intervention scale.

The auxiliary target cap $\lambda_{\max}^{A}$ mainly controls the Base-to-novel generalization balance. As Figure~\ref{fig:hyper_sensitivity}(a) shows, raising the cap from 0.10 to 0.40 improves EuroSAT HM from 90.32 to 92.02, driven by a novel-class gain from 84.03 to 87.01 while base accuracy stays high between 97.12 and 97.64. Raising the cap further to 0.50 reduces HM to 90.95 and novel accuracy to 85.31, so auxiliary intervention is helpful only while it remains bounded. Cross-dataset transfer accuracy is comparatively flat, ranging from 69.02 at 0.10 to a peak of 69.43 at $\lambda_{\max}^{A}=0.30$. This sweep supports the bounded-intervention design. The policy needs enough capacity to correct the teacher, but excessive auxiliary mass weakens the Base-to-novel generalization balance.

Policy regularization prevents unstable target construction, but its best strength is intermediate. As Figure~\ref{fig:hyper_sensitivity}(b) shows, removing regularization ($\alpha_r=0$) lowers EuroSAT HM to 87.71 and novel-class accuracy to 80.21, the signature of an unconstrained policy making overly aggressive target updates. A small regularizer, $\alpha_r=0.01$, restores HM to 90.72, and $\alpha_r=0.05$ gives the best Base-to-novel generalization result (97.64 Base, 87.01 Novel, 92.02 HM). Cross-dataset transfer accuracy peaks at $\alpha_r=0.10$ with 69.43 and stays close at $\alpha_r=0.05$ with 69.31, while raising the regularizer to 0.20 lowers both HM and transfer accuracy. The sweep thus confirms the role of $\alpha_r$ as a stabilizer that should curb erratic actions without suppressing useful adaptation.

The action-strength sweep shows that stronger intervention is not always better. Figure~\ref{fig:hyper_sensitivity}(c) reports the best EuroSAT Base-to-novel generalization HM at $s=0.50$, where Novel reaches 87.01 and HM reaches 92.02. A smaller scale, $s=0.25$, gives the highest Base accuracy of 97.76 but lowers Novel to 82.23 and HM to 89.33. Cross-dataset transfer accuracy aligns less with HM. It falls to 65.25 at $s=1.00$ and recovers to 69.43 at $s=1.50$, a setting whose HM is only 89.63. Action strength therefore controls how sharply the controller translates reliability cues into target changes, and the best setting for Cross-dataset transfer can differ from the best setting for Base-to-novel generalization balance. This reinforces why \ours{} uses bounded actions and separate controls for mixture, weight, and temperature. Taken together, the sweeps show that \ours{} is governed by two forms of control. The auxiliary cap determines how much non-teacher supervision is available (the best Base-to-novel generalization balance appears at $\lambda_{\max}^{A}=0.40$ and the best Cross-dataset transfer accuracy at 0.30), while policy regularization and action scale determine how safely that supervision is used, with $\alpha_r=0.05$ and $s=0.50$ best for Base-to-novel generalization. Cross-dataset transfer accuracy is less monotonic, suggesting that transfer robustness depends on both the amount of intervention and the way the controller applies it. These results support the central design choice of \ours{}. Target construction should be adaptive enough to correct unreliable teacher targets, yet each intervention should remain bounded and feedback-stabilized.

\section{Conclusion}

This paper presented \ours{}, an on-policy distillation framework for vision-language model adaptation. Instead of imposing a fixed teacher-centered target on every training sample, \ours{} treats target construction as a policy decision that adaptively mixes the adapted teacher, the frozen zero-shot prior, and hard-label supervision according to reliability cues and validation feedback. Because the policy only shapes the training objective, the distilled student keeps the same inference architecture and adds no test-time computation. Experiments on Base-to-novel generalization and Cross-dataset transfer show that adaptive target construction improves the robustness and transferability of vision-language distillation, with clear gains on challenging datasets such as FGVCAircraft, DTD, and EuroSAT. Ablation studies further confirm that the gains do not simply come from adding more supervision sources. The zero-shot prior, hard-label anchoring, validation feedback, and reliability-aware policy cues each contribute to stable target construction. These results suggest that effective vision-language distillation requires not only multiple sources of supervision, but also a mechanism that decides when and how each source should influence student learning. Future work can explore more expressive policy-learning objectives, stronger validation-feedback signals, and applications beyond classification.

\bibliographystyle{cas-model2-names}
\bibliography{reference}

\end{document}